\documentclass[letterpaper, 10 pt, conference]{ieeeconf}  

\IEEEoverridecommandlockouts                              

\usepackage{array}
\usepackage[T1]{fontenc}
\usepackage{upquote}
\usepackage{xcolor}
\usepackage{graphicx}
\usepackage{amsmath}
\usepackage{amsfonts}
\usepackage{bm}
\usepackage{booktabs}
\usepackage{color}
\usepackage{url}
\usepackage{amsmath}
\usepackage{algorithm}
\usepackage{algpseudocode}
\usepackage{mdframed}
\usepackage{hyperref}
\usepackage{multirow}
\usepackage{comment}
\usepackage{makecell}

\title{\LARGE \bf VLM-driven Behavior Tree for Context-aware Task Planning}

\author{Naoki Wake$^{1}$, Atsushi Kanehira$^{1}$, Jun Takamatsu$^{1}$, Kazuhiro Sasabuchi$^{1}$, and Katsushi Ikeuchi$^{1}$%
\thanks{$^{1}$Applied Robotics Research,                Microsoft, Redmond, WA, USA}
}

\begin{document}
\maketitle
\thispagestyle{empty}
\pagestyle{empty}

\begin{abstract}
The use of Large Language Models (LLMs) for generating Behavior Trees (BTs) has recently gained attention in the robotics community, yet remains in its early stages of development. In this paper, we propose a novel framework that leverages Vision-Language Models (VLMs) to interactively generate and edit BTs that address visual conditions, enabling context-aware robot operations in visually complex environments. A key feature of our approach lies in the conditional control through self-prompted visual conditions. Specifically, the VLM generates BTs with visual condition nodes, where conditions are expressed as free-form text. Another VLM process integrates the text into its prompt and evaluates the conditions against real-world images during robot execution. We validated our framework in a real-world cafe scenario, demonstrating both its feasibility and limitations. The sample code is available at \href{https://github.com/microsoft/scene-aware-robot-BT-planner}{https://github.com/microsoft/scene-aware-robot-BT-planner}.
\end{abstract}
%
%
%
%
%

\section{Introduction}
Advancements in robotic hardware have accelerated the commercialization of robots across diverse fields such as logistics, manufacturing, and healthcare, while enabling deployment in dynamic and diverse environments beyond traditional factories. This trend highlights the need for rapidly updatable programs tailored to each unique use case, with minimal reliance on skilled programmers.
\begin{figure}[t]
  \centering
  \includegraphics[width=\columnwidth]{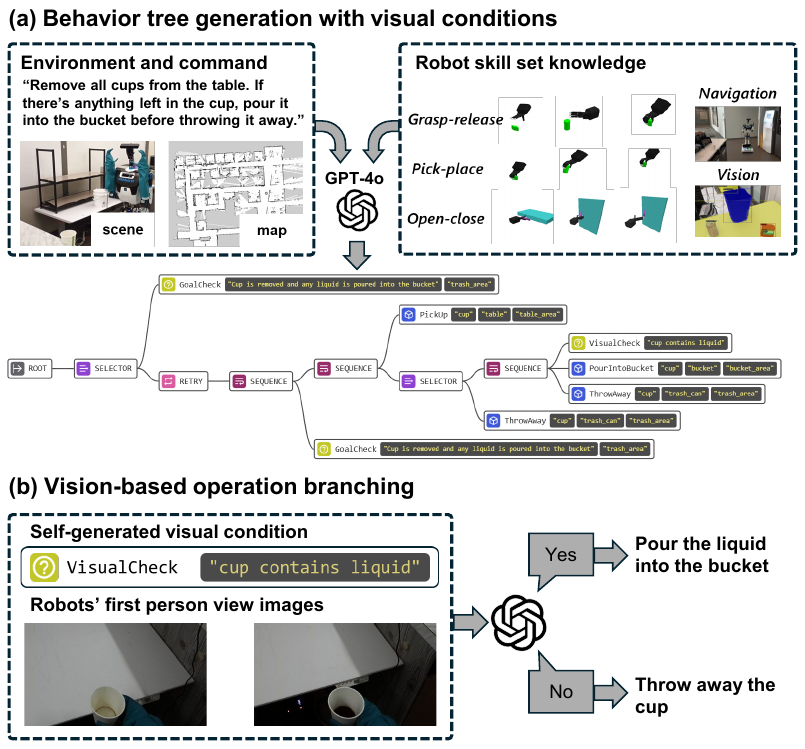}
  \caption{We propose a user-friendly robot system that enables domain experts to program robots interactively. The system takes user instructions, scene details, map information, and the robot's skill set as inputs,  and converts them into a visual program represented as a Behavior Tree (BT). The BT incorporates visual condition nodes that dynamically switch the robot's behavior based on real-world images during execution.}
  \label{fig:topfigure}
\end{figure}
Behavior Trees (BT) have emerged as a potential solution to address the need as they offer modularity, reusability, and readability in robotic programming~\cite{iovino2022survey,colledanchise2021implementation,ogren2022behavior}. Existing BT generation methods, such as those based on genetic programming~\cite{styrud2022combining,colledanchise2018learning}, have proven effective but are typically limited to specific scenarios. With the advent of Large Language Models (LLMs), new approaches have aimed to leverage LLMs' superior language comprehension capabilities to generate diverse BTs~\cite{wang2024mosaic,zhou2024llm,gonzalez2024boosting,chen2024integrating,chen2024efficient,izzo2024btgenbot,li2024study,lykov2023llmbrain,ao2024llm,cao2023robot,lykov2023llm}, including conditional branches~\cite{yang2023robot,styrud2024automatic}. However, this field remains in its early stages of development.

In this paper, we propose a novel BT generation framework that leverages Vision-Language Models (VLMs) to implement ``self-prompted visual conditions (Fig.~\ref{fig:topfigure}).'' While existing LLM-based approaches primarily focus on BT generation, our method expands the scope to include runtime switching between BT nodes based on visual condition evaluation. Specifically, the VLM integrates flexible condition nodes into BTs. These nodes include conditional statements and are evaluated against real-world images during robot execution. For example, given an instruction like ``remove the cups from the table,'' the VLM generates a condition node to check if the table is clear of cups, and includes it in the BT. This condition (i.e., ``The table is clear of cups.'') is evaluated against real-time visual inputs, thereby enabling the VLMs to make environment-aware decisions. Furthermore, as highlighted in previous research on LLM-driven BT generation, ensuring program safety and transparency is essential. To address this, we developed an interface that integrates BT visualization with interactive editing, enhancing both safety and transparency in the generated BTs.

We validated the proposed framework using a humanoid robot in a real-world cafe scenario. Tasks involving conditional branches, such as ``clear all cups from the table'' and ``if a cup contains liquid, discard the contents before placing it in the trash; otherwise, place it directly in the trash,'' were generated as BTs and tested on the robot. The results demonstrated that our method effectively resolved the specified tasks.

The contributions of this paper are: 
\begin{itemize}
    \item We propose a novel BT generation framework that incorporates ``self-prompted visual conditions'' using VLMs.
    \item We develop an interface for BT visualization and interactive editing, enabling flexible and safe task generation.
    \item We validate the proposed method on a real-world robot and demonstrate its effectiveness. Additionally, we release the source code and guidelines to support the robotics community.
\end{itemize}

\section{Related Work}
\subsection{LLM-based BT generation}



While BT generation methods have traditionally employed learning-based approaches~\cite{li2024study,lykov2023llm,lykov2023llm}, recent work has increasingly focused on utilizing pre-trained LLMs through pre-prompting or fine-tuning. Early studies leveraging pre-trained LLMs often employed BT templates~\cite{cao2023robot}, or used LLMs only to produce preliminary information to help a BT builder~\cite{yang2023robot,zhou2024llm}. 
More recently, however, it has become common to have LLMs directly generate BTs~\cite{izzo2024btgenbot,mower2024ros,chen2024integrating,gonzalez2024boosting}. This shift is likely driven by the improved capabilities of LLMs in handling longer contexts and larger token windows.

While our study aligns with the approach of directly generating BTs, a key distinction is the incorporation of visually-driven conditions. Expressed as free-form text, these conditions serve as prompts to be verified during robot execution. This method enables flexible conditional checks beyond the capabilities of conventional task-specific vision models. While one study proposed utilizing a vision-language model (VLM) for detecting failures~\cite{ahmad2024addressing}, to the best of our knowledge, no prior work has attempted to use VLMs to address conditional branches within BTs. 

\subsection{BT for robotics}
Originating in the field of gaming, BTs provide a state-free, hierarchical structure where execution nodes and control-flow nodes work together to manage complex behaviors~\cite{colledanchise2018behavior}. 
BTs are highly suited for robotics applications owing to their modular tree structure, flexibility for modifications, and high readability~\cite{iovino2022survey,colledanchise2021implementation,ogren2022behavior}. For example, the BehaviorTree.CPP library\footnote{https://www.behaviortree.dev/} has become the de facto standard for behavior tree implementations in the Robot Operating System 2 (ROS2), as a result of its integration within the widely used Navigation2 framework~\cite{macenski2020marathon}.

Ensuring the safety and success of generated BTs is another critical factor in robotics. Some researchers have introduced syntax-checking capabilities through reflective feedback~\cite{chen2024efficient,chen2024integrating} or iterative corrections using a simulator~\cite{ao2024llm}. Nevertheless, reflective methods are costly in terms of time and resource consumption due to the increase in LLM calls. Recent findings suggest that a human-in-the-loop approach can outperform fully automated correction methods~\cite{ao2024llm}. Therefore, we have chosen to adopt a human-in-the-loop strategy to assist in generating safe BTs. To this end, we have developed an interface that integrates interactive BT editing, enabling human users to guide and refine the final outcomes (see Section~\ref{interface}).


\subsection{Use of VLM for robotics}
The advancements in LLMs and VLMs have significantly contributed to progress in robotics applications (\cite{Kovalev2022ApplicationOP,dang2024explainable} provide an overview). While one major research direction is to fine-tune vision-language action models for end-to-end robot control (e.g., ~\cite{kim2024openvla,team2024octo}), a growing number of studies demonstrate the effectiveness of off-the-shelf VLMs for high-level task planning and spatial reasoning (e.g., \cite{wake2024gpt,wake2024open,nasiriany2024pivot}). Following this trend, we propose using an off-the-shelf VLM, specifically GPT-4o, to generate BTs from scene data and instructions, as well as to monitor their execution progress using egocentric vision during robot operations.

\section{VLM-driven BT generation framework}\label{method}
In order to obtain a BT structure, both a pre-prompt and a human instruction are provided to a VLM. The structure and content of the pre-prompt are based on our previous paper~\cite{wake_chatgpt}, as outlined below:
\begin{itemize}
    \item \textbf{Role Prompt}: Explains the general task overview to the VLM.
    \item \textbf{Environment Prompt}: Describes how the input scene information is structured, along with specific details about the scene and map.
    \item \textbf{Output Prompt}: Specifies the expected output from the VLM and its format.
    \item \textbf{Action Prompt}: Defines the list of robot actions at the granularity of BTs and their arguments.
    \item \textbf{Example Prompt}: Provides examples of the output format.
\end{itemize}

These pre-prompts are concatenated in a conversational format. The user's textual instructions are then appended to this pre-prompt to generate output from the VLM. All conversations are temporarily stored in memory during the process so that, when the user provides feedback, modifications can be made based on the recent conversation history.

It is noteworthy that in this prompt, scene and map information are provided as text, which means that BTs can be generated even by an LLM without relying on a VLM. However, for branching decisions that depend on the robot’s egocentric vision, we employ a VLM (i.e., GPT-4o) to maintain a unified workflow under a single model framework.
\subsection{Role Prompt}
This prompt provides the VLM with context for the task. Figure~\ref{fig:p_role} shows an example.

\begin{figure}[ht]
\begin{mdframed}[backgroundcolor=gray!20] 
\begin{flushleft}
\color[rgb]{0.3,0.3,0.3}\normalsize 
You are an excellent interpreter of human instructions for operating a robot. Given instructions, information about the working environment, and details of the actions the robot can perform, you will break them down into a sequence of robot actions represented as a Behavior Tree.
\end{flushleft}
\end{mdframed}
\caption{An example of a role prompt.}
\label{fig:p_role}
\end{figure}

\subsection{Environment prompt}\label{environment}
This prompt explains the format used to represent working environments. The idea of using environmental data for LLM task planning is a widely applied technique, including the study aiming for BT generation~\cite{zhou2024llm,ao2024llm,chen2024efficient}, as it effectively constrains the set of manipulable objects and serves as a valuable frame of reference. Figure~\ref{fig:p_env} shows an example of the definitions. In this example, the \textit{semantic map} contains a list of location names used for robot navigation. The \textit{object metadata} includes a list of objects subject to manipulation. The \textit{assets metadata} contains a list of environment elements that are not directly manipulated but are relevant to the manipulation process (e.g., table). The \textit{asset object relations} describe the spatial relationships between objects and assets (e.g., "on\_something" or "inside\_something"), while the \textit{location asset relations} specify the proximity of assets to specific locations. This set of information has been identified as practically effective for scenarios involving mobile manipulation. However, it is noteworthy that additions or deletions to the information can be made on a case-by-case basis to adapt to specific robotic scenarios. For example, manipulation tasks that do not involve mobile navigation are unlikely to require semantic map information.

Notably, operating real robots requires access to additional parameterized information associated with locations, assets, and objects. For instance, precise coordinates of the locations, the approach direction, and the grasping strategy for manipulating specific objects are essential for generating motor commands. A detailed example of environment information is provided in the Appendix (Fig.~\ref{fig:environment_info}) for reference. These details can be obtained through user demonstrations~\cite{wake2020verbal,wake2020grasp,saito2022task}, VLM-based environment recognition~\cite{wake2024gpt}, or manual input from users or operators. However, the collection of environmental data falls outside the scope of this paper. 

\begin{figure}[ht]
    \begin{mdframed}[backgroundcolor=gray!20] 
    \begin{flushleft}
    \color[rgb]{0.3,0.3,0.3}\normalsize 
    Information about environments and objects is given as a Python dictionary. Example:\\
    \textquotedbl\textquotedbl\textquotedbl\\
    \{\hspace*{1em}\\
    \hspace*{1em}"environment": \{\hspace*{1em}\\
        \hspace*{2em}"semantic\_map\_locations": \{\\
        \hspace*{3em}...information about locations navigable by the robot...\\
        \hspace*{2em}\},\\
        \hspace*{2em}"objects\_metadata": \{\\
        \hspace*{3em}...information about the objects in the scene...\\
        \hspace*{2em}\},\\
        \hspace*{2em}"assets\_metadata": \{\\
        \hspace*{3em}...information about the assets in the scene...\\
        \hspace*{2em}\},\\
        \hspace*{2em}"asset\_object\_relations": \{\\
        \hspace*{3em}...information about the asset-object relations...\\
        \hspace*{2em}\},\\
        \hspace*{2em}"location\_asset\_relations": \{\\
        \hspace*{3em}...information about the location-asset relations...\\
        \hspace*{2em}\}\\
    \hspace*{1em}\}\\
    \}\\
    \textquotedbl\textquotedbl\textquotedbl\\
    Relationships and metadata are represented using structured sets:\\
    \textquotedbl\textquotedbl\textquotedbl\\
    \textbf{State list}\\
    - on\_something(<something>): Object is located on <something>\\
    - inside\_something(<something>): Object is located inside <something>\\
    - inside\_hand(): Object is being grasped by a robot hand\\
    - closed(): Object can be opened\\
    - open(): Object can be closed or kept open\\
    \textquotedbl\textquotedbl\textquotedbl\\
    <something> should be one of the assets or objects in the environment.
    \end{flushleft}
    \end{mdframed}
  \caption{An example of an environment prompt.}
  \label{fig:p_env}
\end{figure}

\subsection{Output prompt}\label{format}
This prompt explains the information expected as output from the VLM and its format. Our preliminary observations indicate that the GPT-4o model possesses basic knowledge of BTs (this is not shown in a figure but can be verified, for example, by querying ``What is a Behavior Tree and explain its concept'' to GPT-4o). However, for clarity, we explicitly defined the structure of BTs and the roles of each node as part of the pre-prompt. While prior research aiming to directly generate BTs from LLMs often adopted an XML format, we opted for JSON in this work as it facilitates ease parsing.

In the prompt example illustrated in Fig.~\ref{fig:bt_prompt}, the system is assumed to repeatedly execute the main sequence until it visually confirms that the ultimate goal has been achieved at a specific location. Under this assumption, the VLM is tasked with determining the following based on the user’s instruction:
\begin{itemize}
    \item main\_sequence: The sequence of operations repeated until the ultimate goal is achieved.
    \item ultimate\_goal: The visual condition defining the successful achievement of the goal.
    \item where\_to\_check\_goal: The location where the ultimate goal is visually confirmed.
\end{itemize}

The template illustrated here was defined for a mobile navigation scenario in a cafe environment, and the use of templates or visually-conditioned final goals is not necessarily applicable in general. The key role of the output prompt lies 
in clarifying the use of the BT method and specifying the information the VLM should include in its output.

\begin{figure}[ht]
    \begin{mdframed}[backgroundcolor=gray!20]
    \begin{flushleft}
    \color[rgb]{0.3,0.3,0.3}\normalsize
    You divide the actions given in the text into detailed robot actions and organize them into a behavior-tree-like format.\\

    \textbf{Behavior tree nodes:}\\
    \textquotedbl\textquotedbl\textquotedbl\\
    Sequence nodes execute all children sequentially and fail if any child fails.\\
    Selector nodes execute children until one succeeds or all fail.\\
    Retry decorators retry their child node until it succeeds.\\
    \textquotedbl\textquotedbl\textquotedbl\\

    \textbf{Your response format:}\\
    \textquotedbl\textquotedbl\textquotedbl\\
    \{\hspace*{1em}\\
    \hspace*{1em}"main\_sequence": action sequence in the BT format,\\
    \hspace*{1em}"ultimate\_goal": desired visual state,\\
    \hspace*{1em}"where\_to\_check\_goal": location to check the goal\\
    \}\\
    \textquotedbl\textquotedbl\textquotedbl\\

    \textbf{Compiled behavior tree:}\\
    \textquotedbl\textquotedbl\textquotedbl\\
    root \{\\
    \hspace*{1em}selector \{\\
    \hspace*{2em}action [GoalCheck, ultimate\_goal, where\_to\_check\_goal]\\
    \hspace*{2em}retry \{\\
    \hspace*{3em}sequence \{\\
    \hspace*{4em}main\_sequence\\
    \hspace*{3em}\}\\
    \hspace*{2em}\}\\
    \hspace*{1em}\}\\
    \}\\
    \textquotedbl\textquotedbl\textquotedbl\\
    \end{flushleft}
    \end{mdframed}
  \caption{An example of an output prompt.}
  \label{fig:bt_prompt}
\end{figure}

\subsection{Action prompt}\label{actionprompt}
This prompt describes a list of robot actions defined at the granularity of BT nodes and their corresponding arguments. Figure~\ref{fig:leaf_nodes} shows an example. In a BT, each node represents a specific action or condition evaluation. Here, the placeholder, node\_definition\_placeholder is replaced with robot actions tailored to each scenario.

Notably, the actions defined in this prompt are independent of the granularity of the smallest unit of robot tasks. Instead, BT nodes can represent a sequence of robot tasks (so-called task cohesion~\cite{yanaokura2022multimodal}). In many cases, the level of granularity at which humans can easily understand or describe actions linguistically is broader than the actual execution units of a robot. For instance, an action such as ``pick up an object'' consists of a series of robotic operations during execution: locating the object, orienting the camera, approaching the object, and grasping it. Defining actions at an excessively fine granularity can hinder visual debugging and make the action sequences highly dependent on specific robot hardware.

Thus, we recommend defining actions at a granularity that is easy to represent, hardware-agnostic, and likely to be reusable across various scenarios. By pre-selecting actions that meet these criteria, this BT-generation framework can ensure both versatility within a specific scenario (e.g., a cafe environment) and independence from robot hardware (see Section~\ref{diverse}). Correspondence between BT nodes and robot action sequences should be defined separately for each robot. 
A detailed breakdown of how each BT node is expanded into the low-level robot actions used in our system is provided in the Appendix (see Table~\ref{tab:bt_node_expansion}).

Examples of the nodes we prepared for the cafe scenario, along with their definitions, are provided in Table~\ref{tab:action_nodes_end2end}. Several nodes include a location argument to account for the robot's base movement. We define two visual check functions: VisualCheck and GoalCheck. VisualCheck is invoked immediately after the execution of the preceding action, without requiring the robot's base movement, to confirm whether certain visual conditions are satisfied. GoalCheck, on the other hand, is primarily used to verify the completion of the entire BT. GoalCheck takes both visual requirements and location information, specifying where these requirements should be confirmed, as arguments. We introduced two distinct visual check functions because verifying the completion of the BT may not always be feasible at the location where the previous action was performed. For example, when clearing all cups from a table, the robot must pick up the cups, dispose of them in a trash bin, and then return to the table to ensure that no cups remain on it.
\begin{figure}[ht]
    \begin{mdframed}[backgroundcolor=gray!20]
    \begin{flushleft}
    \color[rgb]{0.3,0.3,0.3}\normalsize
    Leaf nodes can be split into two types: \\
    \textbf{Actions}: Perform some kind of action. \\
    \textbf{Conditions}: Check some kind of condition. \\

    Necessary and sufficient nodes are defined as follows: \\
    \textquotedbl\textquotedbl\textquotedbl\\
    \textbf{Robot action list}: \\
    \texttt{node\_definition\_placeholder} \\
    \textquotedbl\textquotedbl\textquotedbl
    \end{flushleft}
    \end{mdframed}
    \caption{An example of an action prompt.}
    \label{fig:leaf_nodes}
\end{figure}

\begin{table}[ht]
\centering
\caption{Defined Action and Condition Nodes for the Proposed BT Generator}

\begin{tabular}{|l|p{0.68\columnwidth}|} 
\hline
\textbf{Node Name} & \textbf{Description} \\
\hline
PickUp & Navigate the robot to \texttt{@location}, look at \texttt{@object}, and grasp it from \texttt{@asset}. \\
\hline
PourIntoBucket & Navigate the robot to \texttt{@location}, then tilt \texttt{@object} in its hand to pour the contents into \texttt{@asset}. \\
& This action should only appear after \texttt{@object} has been picked up in a previous action. \\
\hline
ThrowAway & Navigate the robot to \texttt{@location}, then throw \texttt{@object} into \texttt{@asset}. \\
& This action should only appear after \texttt{@object} has been picked up in a previous action. \\
\hline
VisualCheck & Invoke a vision language model. This node returns true if a vision system confirms that \texttt{@true\_situation} is satisfied. \\
\hline
GoalCheck & Verify the completion of the BT based on \texttt{@ultimate\_goal} at the specified location, \texttt{@where\_to\_check\_goal} (see Section~\ref{format}).\\
\hline
\end{tabular}
\label{tab:action_nodes_end2end}
\end{table}

\subsection{Example prompt}
This prompt provides examples of the expected output format. By presenting several representative scenarios, it helps stabilize the output and reduces the effort required for corrections through feedback.

\subsection{Customization for various use cases}
The pre-prompts are examples prepared for the cafe scenario and can be customized for different scenarios, robot hardware, and environments. Table~\ref{tab:customize} summarizes the components that should be adjusted according to the requirements of customization.
\begin{table}[ht]
\centering
\begin{tabular}{|p{0.35\linewidth}|p{0.50\linewidth}|}
\hline
\textbf{Customization aspect} & \textbf{Components to adjust} \\ \hline
Scenario (e.g., cafe, hospital)& Action prompt (Section~\ref{actionprompt} and Table~\ref{tab:action_nodes}) \\ 
\hline
Robot hardware & Mapping of BT nodes to robot action sequences (Table~\ref{tab:bt_node_expansion})\\
\hline
Environment & Environment prompt (Table~\ref{environment} and Fig.~\ref{fig:environment_info})\\ 
\hline
\end{tabular}
\caption{Variables and parts to customize}
\label{tab:customize}
\end{table}

\section{Interactive BT Builder}\label{interface}
\begin{figure}[t]
  \centering
  \includegraphics[width=\columnwidth]{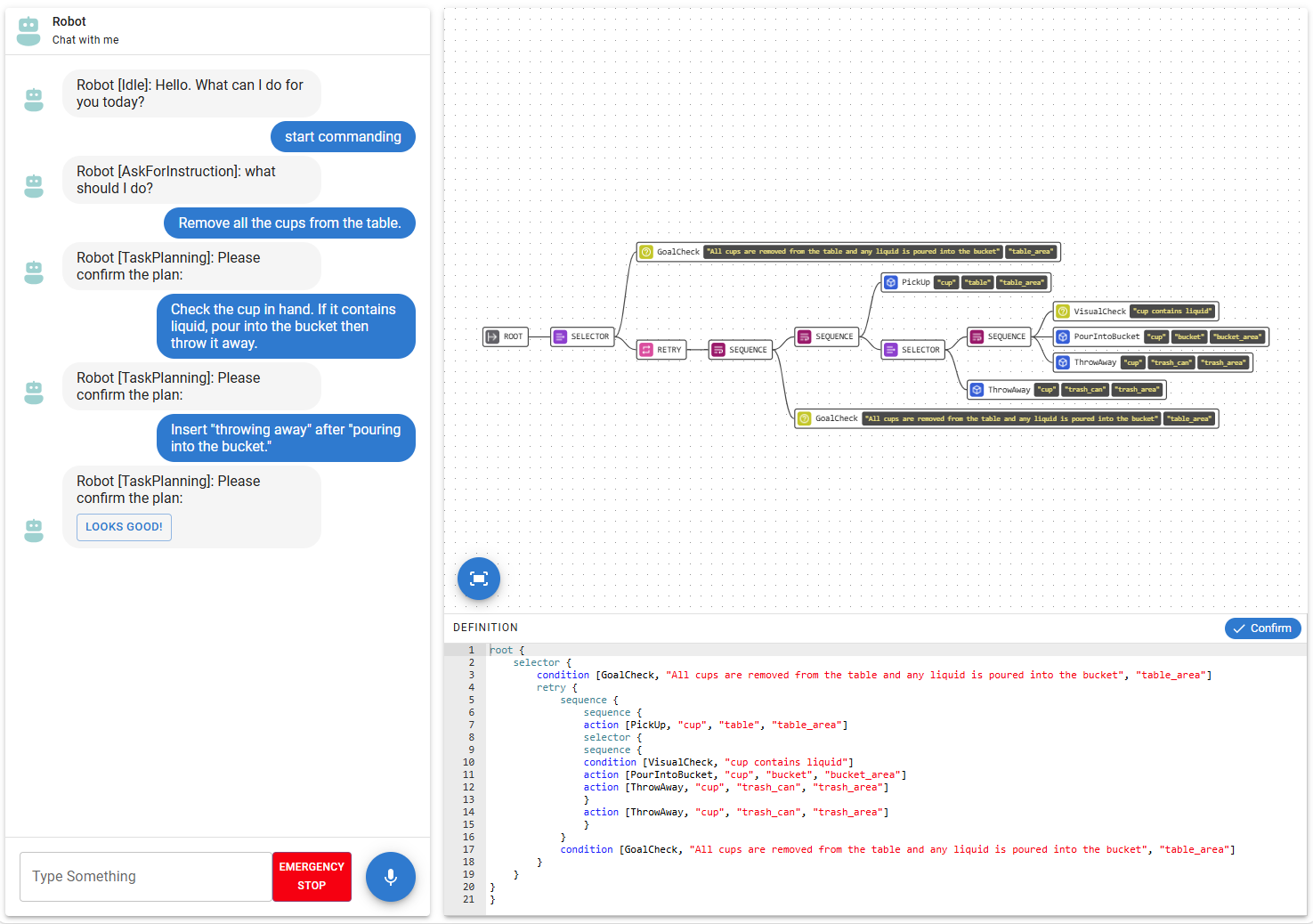}
  \caption{Our developed interactive BT builder. Left panel: A chat window for interactively instructing the VLM. Input can be provided via text or voice, and the VLM's responses are displayed as text and simultaneously played back as audio. Top-right panel: A BT visualization window for the constructed BT. Bottom-right panel: A window for directly editing the BT in text format.}
  \label{fig:webui}
\end{figure}

We developed an interface for interactively constructing BTs using a VLM equipped with the aforementioned pre-prompts (Fig.~\ref{fig:webui}). The BT visualization and text editing components leverage a third-party BT editing library, Mistreevous\footnote{https://github.com/nikkorn/mistreevous}. This web-based interface comprises three components: a chat window, a BT visualization window, and a BT editing window. The chat window enables interactive instructions to the VLM, with inputs provided via text or voice. The BT visualization window displays the BT generated by the VLM, while the BT editing window allows users to modify the BT directly in text format.

\begin{figure}[t]
  \centering
  \includegraphics[width=\columnwidth]{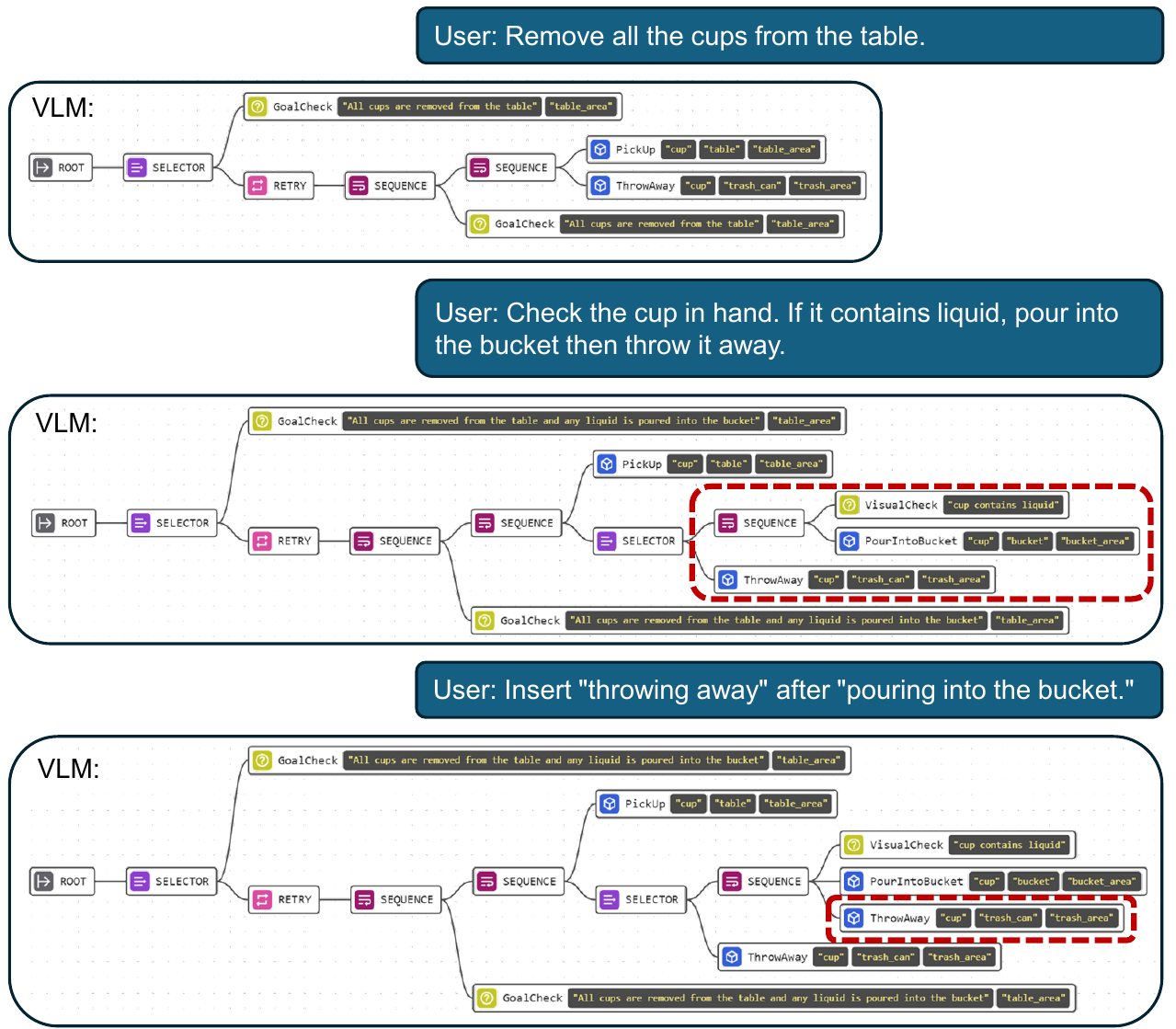}
  \caption{Interactive editing of BT. The red dashed lines represent the tree structures modified by the VLM in response to the most recent user feedback.}
  \label{fig:webui_edit}
\end{figure}

This interface allows users to edit the BT through dialogue without using the editing window, as the VLM retains the conversation history until a correct BT is confirmed by the user. Figure~\ref{fig:webui_edit} illustrates how the VLM modifies the initially generated tree in response to user feedback. 

\section{Experiments}
\subsection{Visual check through self-prompting}
\begin{figure}[t]
  \centering
  \includegraphics[width=\columnwidth]{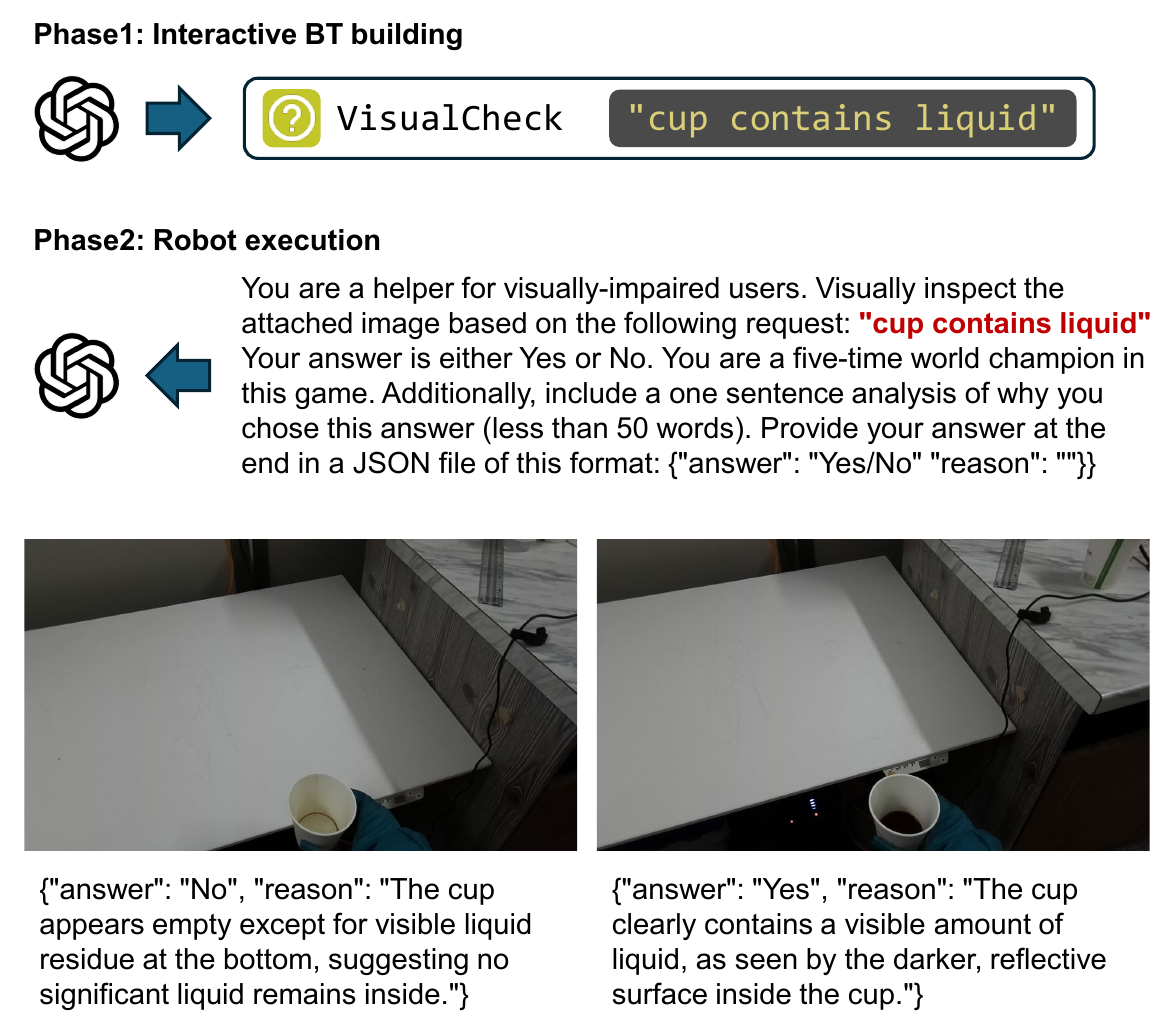}
  \caption{Visual check using VLMs through self-prompting. During the BT building phase, the VLM (e.g., GPT-4o) generates arguments for visual check nodes as free-form language. When the robot reaches these nodes during robot execution, the system internally generates prompts using the arguments and the robot's egocentric image. The next node in the BT is selected based on the VLM's response to these prompts.}
  \label{fig:self_prompt}
\end{figure}

Figure~\ref{fig:self_prompt} shows how the self-prompting method works during a real-robot operation. During the interactive BT building phase, GPT-4o generates arguments for visual check nodes as free-form language. When the robot reaches these nodes during execution, the system internally generates prompts using the arguments and the robot's egocentric image. The next node in the BT is selected based on the VLM's response to these prompts. The bottom images show examples of GPT-4o's responses under different conditions (e.g., whether the cup contains liquid or not), demonstrating the feasibility of dynamically switching the robot's behavior during execution.

\subsection{End-to-end robot experiment}
We conducted experiments applying the proposed BT builder to a real-robot system. In this experiment, we assumed a scenario where a robot works in a cafe and is tasked with disposing of all cups placed on a table into a trash bin. The scenario included uncertain conditions: cups either contained liquid or were empty. If liquid remained in a cup, the robot was required to discard the liquid into a designated container before disposing of the cup into the trash bin. For this experiment, we used a SEED-noid robot (THK)\footnote{https://www.thk.com/jp/en/} equipped with 6-DOF dual arms with one-DOF grippers attached to those.

\begin{figure*}[ht]
  \centering
  \includegraphics[width=\textwidth]{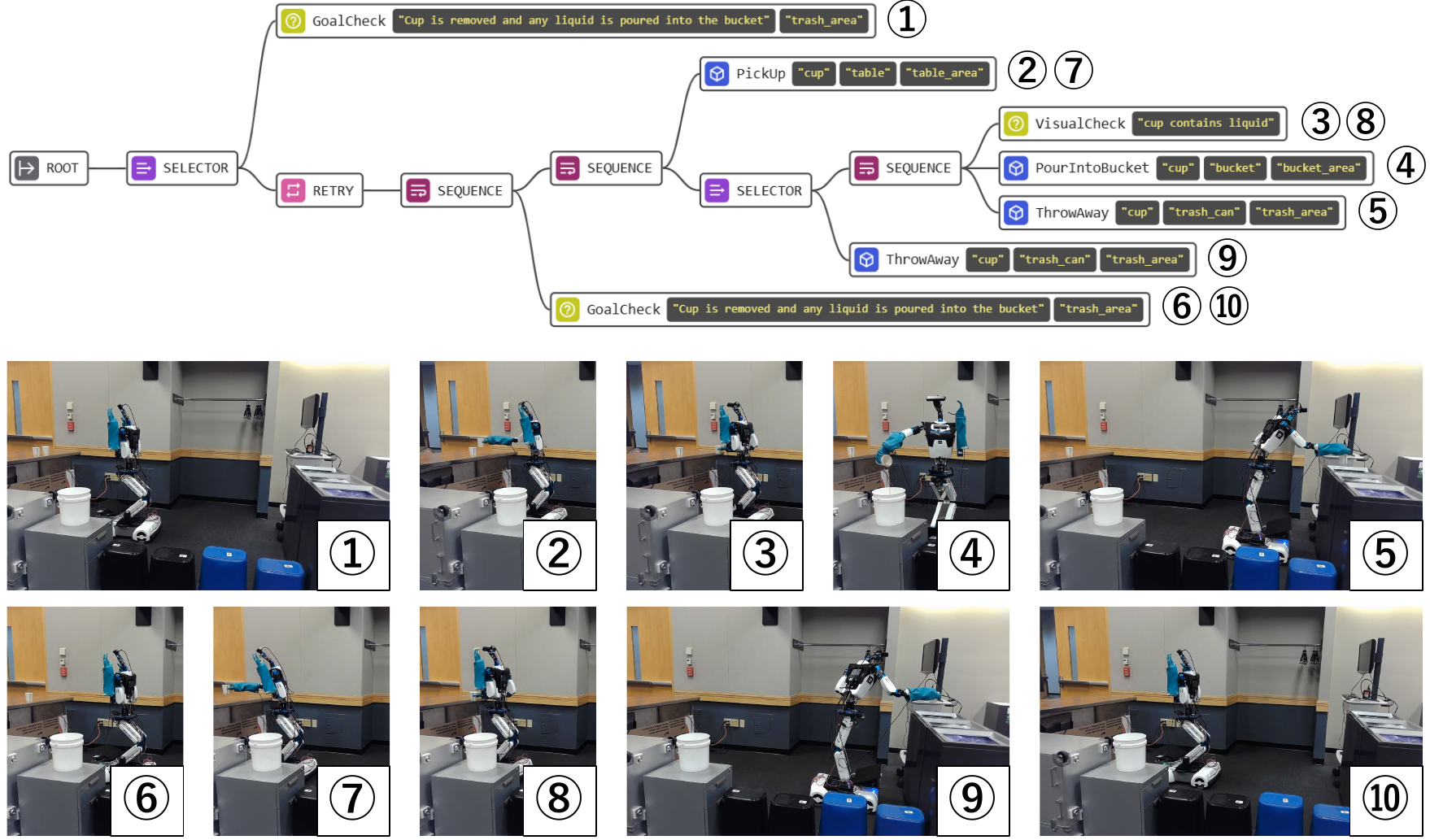}
  \caption{Generated BT and the end-to-end performance of the real robot}
  \label{fig:end2end}
\end{figure*}

The Fig.~\ref{fig:end2end} shows the BT generated by the BT builder and an example of its execution by the robot. During execution, each action node in the BT was internally decomposed into finer-grained robot execution steps such as find, reach, and grasp an object. Robot motions were then generated using a pre-defined environmental map and real-time images captured from the first-person perspective of the robot. As a result, the robot successfully processed cups on the table until none remained, appropriately handling them based on whether they contained liquid. 

We quantitatively evaluated end-to-end performance under a condition where two cups, one containing coffee and one empty, were placed on the table. In this experiment, success was defined as the robot (1) removing all the cups on the table, (2) correctly branching its actions based on the presence or absence of liquid, and (3) ensuring all cups ended up in the trash bin. As shown in Table~\ref{tab:end2end}, the experiment achieved a high success rate of 8/10. The two failure cases occurred when GPT-4o mistakenly recognized an empty cup as containing liquid, which was slightly stained from previously containing coffee, resulting in the robot attempting to discard the liquid into the container. This failure was attributed to the VLM's image recognition performance because the input image had no occlusion and contained the necessary information for a correct judgment to human eyes. 

\begin{table}[ht]
\caption{Success rate at each step of the multi-step manipulation.}
\centering
\resizebox{\columnwidth}{!}{ 
\begin{tabular}{|c|c|c|c|c|}
\hline
\makecell{\textbf{Criteria} \\ \textbf{(Steps)}} & 
\makecell{\textbf{Generating} \\ \textbf{a valid BT}} & 
\makecell{\textbf{Removing} \\ \textbf{all cups}} & 
\makecell{\textbf{Correct} \\ \textbf{branching}} & 
\makecell{\textbf{All cups in} \\ \textbf{the trash bin}} \\ \hline
\textbf{Trials} & 
\makecell{10/10} & 
\makecell{10/10} & 
\makecell{8/10} & 
\makecell{10/10} \\ \hline
\end{tabular}
}
\label{tab:end2end}
\end{table}

\subsection{Building diverse BTs in cafe scenes}~\label{diverse}
\begin{figure*}[t]
  \centering
  \includegraphics[width=0.9\textwidth]{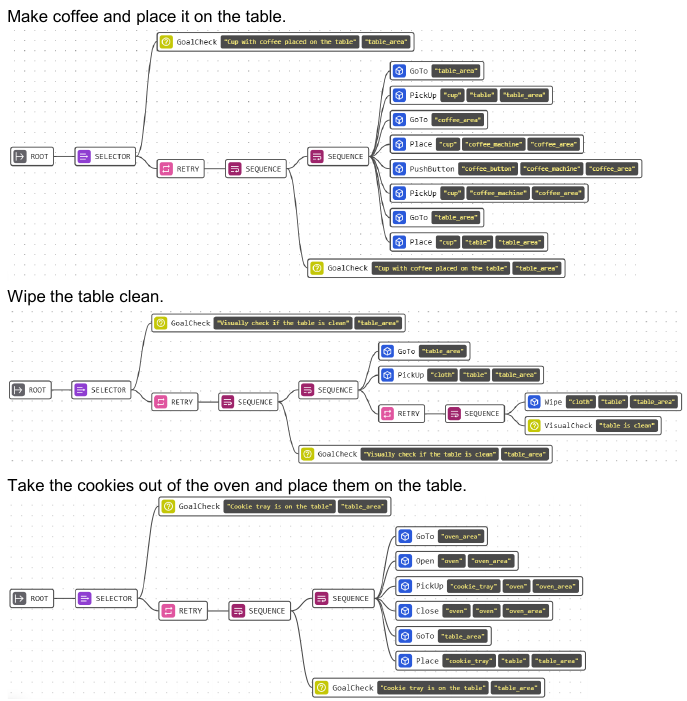}
  \caption{Generation of diverse BTs given a set of action nodes.}
  \label{fig:BT_variation}
\end{figure*}

We qualitatively evaluated the proposed BT builder's applicability to various scenarios. With action nodes defined in Table~\ref{tab:action_nodes} three scenarios were tested: making coffee, wiping a table clean, retrieving cookies from an oven. Figure~\ref{fig:BT_variation} shows the resulting BTs, indicating that the proposed BT builder can adapt to diverse scenarios when provided with an appropriate set of action nodes.

\begin{table}[ht]
\centering
\caption{Defined Action Nodes for the Proposed BT Generator}
\begin{tabular}{|l|p{0.7\columnwidth}|}
\hline
\textbf{Node Name} & \textbf{Description} \\
\hline
GoTo & Navigate the robot to \texttt{@location}. \\
\hline
PickUp & The robot grasps \texttt{@object} from \texttt{@asset}. \\
& This action can only occur after the robot is at \texttt{@location} in a previous action. \\
\hline
Place & The robot places \texttt{@object} in its hand onto \texttt{@asset}. \\
& This action can only occur after \texttt{@object} has been picked up and the robot is at \texttt{@location}. \\
\hline
Pour & The robot tilts \texttt{@object} in its hand to pour the contents into \texttt{@asset}. \\
& This action can only occur after \texttt{@object} has been picked up and the robot is at \texttt{@location}. \\
\hline
Wipe & The robot wipes \texttt{@asset} using \texttt{@object} in its hand. \\
& This action can only occur after \texttt{@object} has been picked up and the robot is at \texttt{@location}. \\
\hline
PushButton & The robot pushes the button on \texttt{@asset}. \\
& \texttt{@object} specifies the button name. This action can only occur after the robot is at \texttt{@location}. \\
\hline
ThrowAway & The robot throws \texttt{@object} into \texttt{@asset}. \\
& This action can only occur after \texttt{@object} has been picked up and the robot is at \texttt{@location}. \\
\hline
Open & The robot grasps a handle on \texttt{@asset} to open it, then releases the handle. \\
& This action can only occur after the robot is at \texttt{@location}. \\
\hline
Close & The robot grasps \texttt{@object} on \texttt{@asset} to close it, then releases \texttt{@object}. \\
& This action can only occur after the robot is at \texttt{@location}. \\
\hline
VisualCheck & Invoke a vision language model. This node returns true if a vision system confirms that \texttt{@true\_situation} is satisfied. \\
\hline
GoalCheck & Verify the completion of the BT based on the ULTIMATE GOAL at the specified location, WHERE
 TO CHECK GOAL (see Section~\ref{format}).\\
\hline
\end{tabular}
\label{tab:action_nodes}
\end{table}

\section{Discussion and conclusion}


This paper proposes a BT builder designed to support domain experts in developing robot programs tailored to on-site requirements. Among existing robot-oriented BT frameworks, this study is unique in leveraging the language processing capabilities of VLMs, while introducing vision-based branching guided by self-defined conditions. This approach enables vision-based conditional handling during execution, allowing for context-aware robot control that reflects user intent through visual conditions. We presented prompting strategies and developed an interface that integrates visualization and user feedback into the BT creation process. Experimental results demonstrated that the proposed system can generate diverse BTs for specific scenarios and that the VLM effectively utilizes self-prompts for decision-making based on the robot's egocentric images. Furthermore, end-to-end experiments with a physical mobile robot validated the feasibility of the system in a cafe scenario.

One limitation of this study is that we did not cover BT generation across arbitrary granularities of robotic actions. We assume that differences in granularity can be addressed by using action nodes with a granularity level similar to those in Table~\ref{tab:action_nodes_end2end}, and by remapping the nodes into the granularities designed for each specific robot. Furthermore, the success rate of visual conditions depends on the performance of the VLM (e.g., GPT-4o). However, this limitation can be mitigated by combining the system with additional object recognition models to enhance performance~\cite{wake2024gpt}.

We hope that this research contributes to advancing frameworks for robot BT generation.

\bibliographystyle{unsrt}
\bibliography{bib}
\appendix

\begin{figure}[ht]
    \begin{mdframed}[backgroundcolor=gray!20]
    \begin{flushleft}
    \color[rgb]{0.3,0.3,0.3}\normalsize
    \{\\
    \hspace*{1em}"semantic\_map\_locations": \{\\
    \hspace*{2em}"table\_area": \{"position": [0.2, 0.2, 0.0], "orientation": [0, 0, -0.04, 1.0]\},\\
    \hspace*{2em}"bucket\_area": \{...\},\\
    \hspace*{2em}"trash\_area": \{...\}\\
    \hspace*{1em}\},\\
    \hspace*{1em}"objects\_metadata": \{\\
    \hspace*{2em}"cup": \{\\
    \hspace*{3em}"right situation": \{\\
    \hspace*{4em}"grasp\_type": "power", \\
    \hspace*{4em}"hand\_laterality": "right",\\
    \hspace*{4em}"approach\_direction": [0.1, 0.5, 0.0],\\
    \hspace*{4em}"depart\_direction": [-0.1, 0.0, 0.0],\\
    \hspace*{4em}"bring\_orientation": [0, 0, -0.6, 0.8]\\
    \hspace*{3em}\}\\
    \hspace*{2em}\}\\
    \hspace*{1em}\},\\
    \hspace*{1em}"assets\_metadata": \{\\
    \hspace*{2em}"table", \\
    \hspace*{2em}"trash\_can",\\
    \hspace*{2em}"bucket"\\
    \hspace*{1em}\},\\
    \hspace*{1em}"asset\_object\_relations": \{\\
    \hspace*{2em}"table": ["(on\_something)cup"],\\
    \hspace*{2em}"trash\_can": [],\\
    \hspace*{2em}"bucket": []\\
    \hspace*{1em}\},\\
    \hspace*{1em}"location\_asset\_relations": \{\\
    \hspace*{2em}"table\_area": ["table"],\\
    \hspace*{2em}"trash\_area": ["trash\_can"],\\
    \hspace*{2em}"bucket\_area": ["bucket"]\\
    \hspace*{1em}\}\\
    \}\\
    \end{flushleft}
    \end{mdframed}
    \caption{
    Example of environment information including semantic map locations, object metadata, assets metadata, and their spatial relationships. Some structures (e.g., \textit{bucket\_area}, \textit{trash\_area}) are abbreviated for brevity as they follow a similar pattern to \textit{table\_area}.
    }
    \label{fig:environment_info}
\end{figure}

\begin{table}[ht]
    \centering
    \caption{Mapping of BT nodes to robot action sequences}
    \begin{tabular}{|p{0.22\linewidth}|p{0.65\linewidth}|}
    \hline
    \textbf{Node Name} & \textbf{Low-level Sub-actions (Sequence)} \\
    \hline

    PickUp &
    \begin{itemize}
        \item \texttt{FIND} (locate the target object)
        \item \texttt{CONDITION} (check if object was found)
        \item \texttt{LOOK} (point cameras at the found object)
        \item \texttt{NAVIGATION} (position the hand near the object)
        \item \texttt{BRING} (adjust the hand position for grasping)
        \item \texttt{LOOK} (re-check object position)
        \item \texttt{GRASP} (grip the object)
        \item \texttt{PICK} (lift the object from its asset or surface)
        \item \texttt{BRING} (adjust the hand position for the next action)
    \end{itemize} \\
    \hline

    PourIntoBucket &
    \begin{itemize}
        \item \texttt{NAVIGATION} (move robot base to the specified location)
        \item \texttt{FIND} (locate the target bucket or asset)
        \item \texttt{LOOK} (align sensors with the bucket position)
        \item \texttt{BRING} (move the hand above the bucket)
        \item \texttt{POUR} (perform the pouring action between specified angles)
        \item \texttt{WAIT} (ensure completion of pouring)
        \item \texttt{POUR} (return the pour angle or re-adjust)
        \item \texttt{BRING} (adjust the hand position)
        \item \texttt{NAVIGATION} (optional step to move base away after pouring)
    \end{itemize} \\
    \hline
    ThrowAway &
    \begin{itemize}
        \item \texttt{NAVIGATION} (move base to the specified location)
        \item \texttt{FIND} (locate the target trash bin or related asset)
        \item \texttt{LOOK} (point sensors at the trash bin)
        \item \texttt{BRING} (adjust arm pose for the throw action)
        \item \texttt{BRING} (further adjust pose, if required)
        \item \texttt{RELEASE} (release the object from the robot's gripper)
        \item \texttt{BRING} (retract arm to a safe position)
    \end{itemize} \\
    \hline
    VisualCheck &
    \begin{itemize}
        \item \texttt{PERCEPTION} (invoke a VLM)
        \item \texttt{CONDITION} (evaluate whether the perception check succeeded)
    \end{itemize} \\
    \hline
    GoalCheck &
    \begin{itemize}
        \item \texttt{PREPARE} (setup for navigation)
        \item \texttt{NAVIGATION} (move base to specified location)
        \item \texttt{PERCEPTION} (invoke a VLM)
        \item \texttt{CONDITION} (evaluate whether the visual check succeeded)
    \end{itemize} \\
    \hline
    \end{tabular}
    \label{tab:bt_node_expansion}
\end{table}

\end{document}